\documentclass[journal]{IEEEtran}
\usepackage{times}
\usepackage{soul}
\usepackage{url}
\usepackage[hidelinks]{hyperref}
\usepackage[utf8]{inputenc}
\usepackage[small]{caption}
\usepackage{graphicx}
\usepackage{amsmath}\allowdisplaybreaks
\usepackage{booktabs}
\usepackage{algorithm}
\usepackage{algorithmic}
\usepackage{amsfonts}
\usepackage{subcaption}
\usepackage{multirow}
\usepackage{epsfig,epstopdf}
\usepackage{bm}
\usepackage[framemethod=tikz]{mdframed}
\usepackage[flushleft]{threeparttable}\usepackage[framemethod=tikz]{mdframed}
\usepackage[flushleft]{threeparttable}
\newtheorem{proof*}{Proof}

\ifCLASSINFOpdf
\else
\fi
\begin{document}
%
\title{Learning Sampling Policy for Faster Derivative Free Optimization}
%
%
%

\author{Zhou Zhai, Bin Gu, and  Heng Huang
	\IEEEcompsocitemizethanks{
		\IEEEcompsocthanksitem
		Z. Zhai  is  with School of Computer \& Software, Nanjing University of Information Science \& Technology, Nanjing, P.R.China (e-mail: zhouzhai@nuist.edu.cn).
		\IEEEcompsocthanksitem
		B. Gu is with the department ofmachine  learning,  Mohamed  bin  Zayed  University  of  Artificial  Intelli-gence, UAE, and with JD Finance America Corporation, Mountain View,CA 94043 USA (jsgubin@gmail.com).
		\IEEEcompsocthanksitem  H.  Huang  is  with  Department  of  Electrical  \&  Computer Engineering, University of Pittsburgh, USA, and with JD Finance America Corporation (e-mail: heng.huang@pitt.edu).
	}
}

%
%

\markboth{Journal of \LaTeX\ Class Files,~Vol.~14, No.~8, August~2015}%
{Shell \MakeLowercase{\textit{et al.}}: Bare Demo of IEEEtran.cls for IEEE Journals}
%



\maketitle

\begin{abstract}	
	Zeroth-order (ZO, also known as derivative-free) methods, which estimate  the gradient only by two function evaluations,  have attracted much attention recently because of its broad applications in machine learning community. The two function evaluations are normally generated with random perturbations from standard Gaussian distribution. To speed up ZO methods, many methods, such as variance reduced stochastic ZO gradients and learning an adaptive Gaussian distribution, have recently been proposed to reduce the variances of ZO gradients. However, it is still an open problem whether there is a space to further improve the convergence of ZO methods. To explore this problem, in this paper, we propose a new reinforcement learning based ZO algorithm (ZO-RL) with learning the sampling policy for generating the perturbations in ZO optimization instead of using random sampling. To find the optimal policy, an actor-critic RL algorithm called deep deterministic policy gradient (DDPG) with two neural network function approximators is adopted. The learned sampling policy guides the  perturbed points in the parameter space to estimate a more accurate ZO gradient. To the best of our knowledge, our ZO-RL is the first algorithm to learn the sampling policy using reinforcement learning for ZO optimization which is parallel to the existing methods. Especially, our ZO-RL can be combined with existing  ZO algorithms that could further accelerate the algorithms. Experimental results for different ZO optimization problems show that our ZO-RL algorithm can effectively reduce the variances of ZO gradient by learning a sampling policy, and converge faster than existing ZO algorithms in different scenarios.
\end{abstract}

\section{Introduction}
Gradient based optimization is an important problem in machine learning.
However, in many fields of science and engineering, explicit gradient information is difficult or even infeasible to obtain.
Zeroth-order (ZO,  also known as derivative-free) optimization has attracted an increasing amount of attention, where the optimizer is provided with only function values (zeroth-order information) instead of explicit gradients (first-order information).
Specifically, the ZO optimization algorithms  first generate  perturbed vectors from a (standard) Gaussian distribution. Based on the sampled perturbed vectors, they query the corresponding function values. Then, they can approximate the gradient information based on the technique of finite difference \cite{chen2019zo}.
ZO optimization can theoretically address a wide range of objectives and has been studied in a large number of fields such as optimization, online learning and bioinformatics \cite{koch2018autotune,mania2018simple,vemula2019contrasting}. 
One of the most important applications of ZO optimization is to generate prediction-evasive adversarial examples in the black-box setting \cite{liu2018stochastic,papernot2017practical}, e.g., crafted images with imperceptible perturbations to deceive a well-trained image classifier into misclassification.

\begin{figure*}[!t]
	\centering
	\centering
	\includegraphics[width=6in]{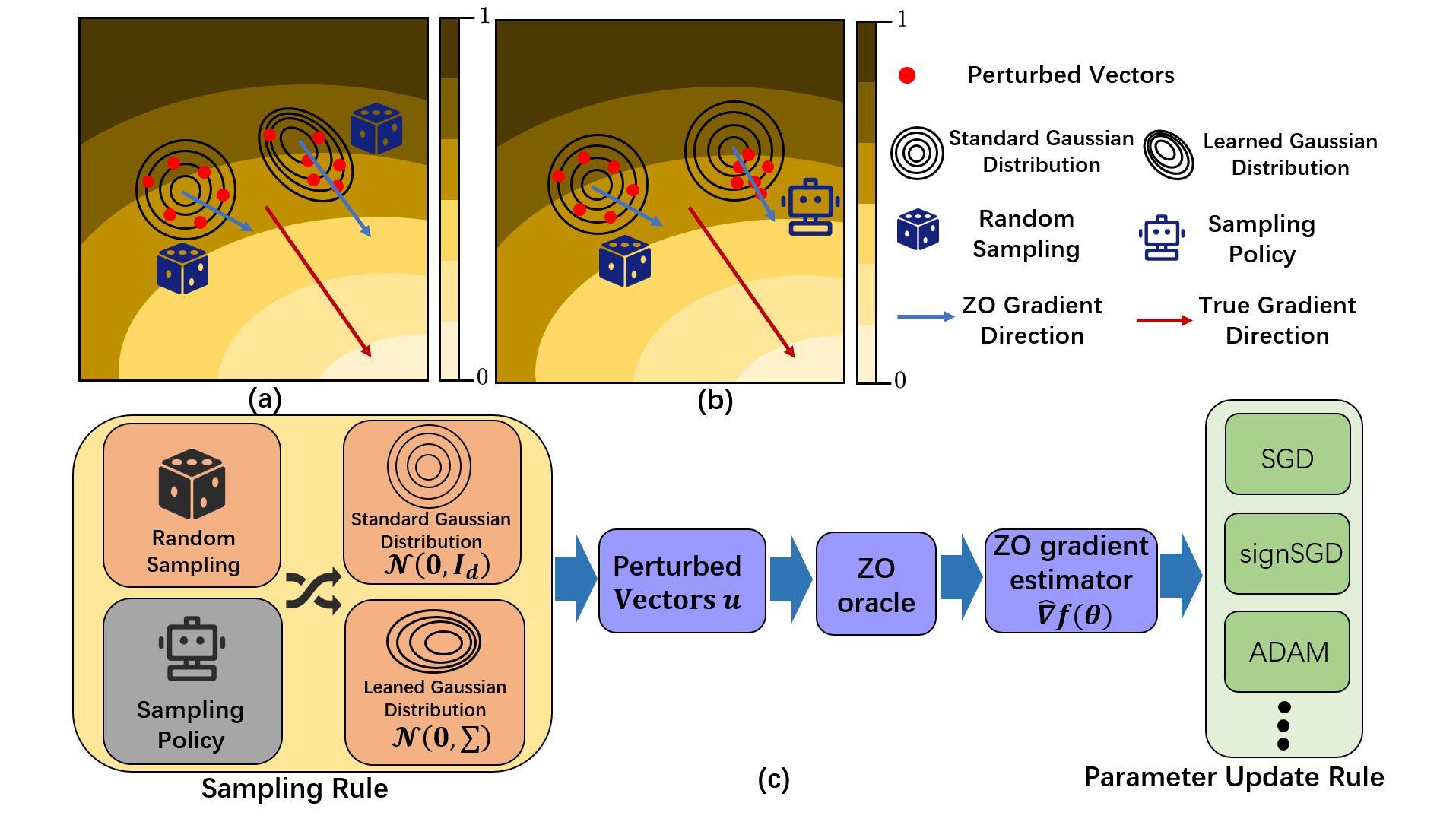}
	\caption{(a) Comparison of the ZO gradient directions obtained by sampling perturbed vectors from the standard Gaussian distribution and a learned Gaussian distribution. (b) Comparison of the ZO gradient directions obtained by a random sampling policy and a learned  sampling policy. (c) The architecture of ZO optimizer. We classify the existing ZO optimization algorithms into two categories (i.e., the sampling rule and the parameter update rule). For sampling rule, we can learn the rules for sampling  distribution and policy. Learning the sampling policy is our main contribution in this paper and it is parallel to existing techniques. Importantly, our learned  sampling policy can be combined with   existing techniques to further accelerate  ZO optimization. 
	}
	\label{pic1}
\end{figure*}

As mentioned above,  standard ZO algorithm constructs a pseudo-gradient by uniformly and randomly sampling some perturbed directions from the standard Gaussian distribution \cite{nesterov2017random,duchi2012randomized}.
However, ZO algorithms often suffer from high variances of ZO gradient estimators, and in turn, hampered convergence rates.
In the past few decades, many ZO methods have been proposed to overcome this problem to speed up the convergence of ZO optimization, which can be divided into two directions.
The first direction is to borrow the gradient descent method used in the first-order algorithm  to improve the parameter update rule in ZO optimization \cite{lian2016comprehensive,chen2017zoo,liu2018signsgd}.
For example, \cite{liu2018zeroth} proposed two accelerated versions of ZO-SVRG, utilizing stochastic variance reduced gradient (SVRG) estimators to reduce the variance.
\cite{liu2018signsgd} extended sign-based stochastic gradient descent (signSGD) method to ZO optimization, compressing the gradient with a single bit by sign operation to mitigate the negative effects of the extreme components of the gradient noise and speed up convergence rate.
\cite{liu2018stochastic} extended SVRG under Gaussian smoothing to reduce the variances of both random samples and random query directions.

The second direction is to utilize the learned sampling distribution to sample perturbed vectors.
Different from sampling perturbed vectors from a standard Gaussian distribution $\mathcal{N}(0,I_d)$, they consider generating perturbed vectors with non-isotropic Gaussian distribution $\mathcal{N}(0,\Sigma)$ such that the covariance matrix $\Sigma$ may not be a scale of the identity matrix.
For example, in a black-box adversarial attack task, there is usually a well-defined significant subspace that can be successfully attacked, and perturbed directions through this subspace leads to a faster convergence.
Similar to this, the evolutionary strategies (ES) such as Natural ES \cite{wierstra2008natural}, CMA-ES \cite{hansen2006cma}, and Guided ES \cite{maheswaranathan2019guided} are proposed to guide the estimation of a descent direction that then can  be passed to the ZO optimizer.
\cite{ruan2019learning} proposed the learning to learn framework to adaptively modify the search distribution with learned recurrent neural networks.

As discussed above, the existing ZO algorithms based on the parameter update rule (i.e., the first direction) can reduce the variance of ZO gradient, while they normally lead to a higher query complexity. The ZO algorithms based on the learned  sampling distribution (i.e., the second direction) also can reduce the variance of of ZO gradient, but they only care about the sampling distribution, and ignore the importance of sampling policy.
We provide Fig \ref{pic1}.(a) to intuitively show the benefit of using the learned sampling distribution, and also provide Fig \ref{pic1}.(b) to  intuitively demonstrate the benefit of using the learned sampling policy.
Since learning the sampling distribution  has been proved that it can improve the convergence  of ZO optimization, naturally we have the following question:
\begin{mdframed}
		Can we learn a sampling policy via using the generating perturbed vectors, instead of simply using random sampling policy, to  further speed up the convergence  of ZO optimization?
\end{mdframed}
In this paper, we provide a positive answer to this question, meanwhile provide a feasible solution to achieve this goal.

Specifically, to solve this challenging problem, we propose a reinforcement learning base zeroth-order algorithm (ZO-RL) to learn the sampling policy in ZO optimization.
Reinforcement learning (RL) is a self-adaptive model and has achieved appealing achievements in practical applications, including playing the ``Atari games” \cite{mnih2015human}, defeating professional Go players \cite{silver2016mastering}, and 3D manipulation of robots \cite{gu2017deep}.
The policy gradient method is a frequently used algorithm in RL.
Model-free policy gradient algorithms are divided into deterministic and stochastic policies.
Compared with the stochastic policy gradient algorithm, the deterministic policy has the advantages of requiring less data to be sampled and high efficiency of the algorithm, and performs stably in a series of tasks with continuous action space.
Thus, to find the optimal policy, an actor-critic RL algorithm called deep deterministic policy gradient (DDPG) \cite{lillicrap2015continuous} with two neural network function approximators is adopted.
The learned sampling policy of RL guide the optimizer to estimate more accurate ZO gradients in the parameter space.
Especially, we can combine our ZO-RL with existing the ZO algorithms that utilize the improved parameter update rule and learned sampling distribution.  
Experimental results for different ZO optimization problems show that our ZO-RL algorithm can effectively reduce the variances of ZO gradient by learning the sampling policy, and converge faster than existing ZO algorithms in different scenarios.

\textbf{Contributions.} The main contributions of this paper are summarized as follows.
\begin{enumerate}
	\item We propose to learn the sampling policy by reinforcement learning instead of using random sampling policy  as in the traditional ZO algorithms to generate perturbed vectors. To the best of our knowledge, our ZO-RL is the first algorithm to learn the sampling policy for ZO optimization which is parallel to the existing ZO methods.
	\item We conduct extensive experiments to show that our ZO-RL algorithm can effectively reduce the variances of ZO gradients by learning a sampling policy, and converge faster than existing ZO algorithms in different scenarios.
\end{enumerate}

\section{Preliminaries}
In this section, we give a brief review of  zeroth-order optimization and  reinforcement learning respectively.
\subsection{Zeroth-Order Optimization}
ZO optimization is widely used in the environments where gradient information is difficult or even impossible to obtain.
For the loss function $f$ with its parameter $x$, we can obtain its ZO gradient estimator by:
\begin{eqnarray}
\hat{\nabla} f(x)= \frac{1}{\mu q}\sum_{i=1}^{q}[f(x+\mu u_i)-f(x)]u_i
\end{eqnarray}
where $\mu>0$ is the smoothing parameter, $\{u_i\}$ are the random query directions drawn from the standard Gaussian distribution, and $q$ is the number of sampled query directions.
We summarized the standard ZO optimization algorithm in Algorithm \ref{alg1}.
Fig \ref{pic1}.(c) shows the  architecture of ZO optimizer. The perturbed vectors (query directions) are first generated by the sampling rule, then passed to ZO Oracle to calculate the ZO gradient estimator. Based on the ZO gradients, we can update solution based on the parameter update rule.

The high variance of  ZO gradient estimators hinders the convergence speed of  ZO algorithms due to the random sampling perturbed vectors.
Thus, the choice of sampling policy and sampling distribution determines the performance of the ZO optimization algorithm.
There has been a research trend to propose sampling the perturbed vectors from some non-isotropic Gaussian distribution instead of  standard Gaussian sampling with an isotropic covariance for random query directions.
They consider sampling perturbed vectors by $u_i\sim \mathcal{N}(0,\Sigma)$ such that the covariance $\Sigma$ may not be a scale of the identity matrix \cite{ye2018hessian,ruan2019learning,maheswaranathan2019guided}.
By learning the significant sampling distribution, more accurate gradient estimators can be obtained for a fixed query budget, which can improve the convergence rate of the ZO optimization task.
However, improving the performance of ZO optimization algorithms by learning sampling policy is still a vacancy in the literature. 
To overcome this problem, in this paper, we use a policy search approach in reinforcement learning  to learn a  sampling policy, and then use it to generate  perturbed vectors to obtain  more accurate gradient estimators, instead of plainly using the random sampling policy to generate  perturbed vectors.

\begin{algorithm}[H]
	\caption{Zeroth-Order (ZO) Optimization Algorithm} \label{alg1}
	\renewcommand{\algorithmicrequire}{\textbf{Input:}}
	\renewcommand{\algorithmicensure}{\textbf{Output:}}
	\begin{algorithmic}[1]
		\REQUIRE Hyper-parameter $\mu$, $q$ and $\eta$.
		\ENSURE $x\in\mathbb{R}^d$ 
		\FOR {$k=0$ to $K-1$} 
		\STATE Sampling $q$ perturbed vectors from the standard Gaussian distribution $u_i\sim\mathcal{N}(0,I_d)$.
		\STATE Calculating the ZO gradient estimator $\hat{\nabla} f(x_t)$.
		\STATE Obtain the next update $x_{k+1}=x_k-\eta\hat{\nabla} f(x_t)$.
		\ENDFOR
	\end{algorithmic}
\end{algorithm}

\subsection{Reinforcement Learning}
RL can be modeled as a Markov decision process (MDP) with a four-tuple of $(S,A,P,R)$, where $S$ means a set of states, $A$ denotes a set of actions, $P$ represents a transition probability function $p(s_{t+1}|s_t, a_t)$, and $R$ is a reward function $R: S\times A\rightarrow R$.
An episode of task denotes that an agent and an environment interact with each other at discrete time steps $t=0, 1, 2, \cdots, T$. 
The agent chooses an action $a_t\in A$ according to its policy under the state $s_t\in S$ of the environment.
If the agent takes a certain action $a_t$, the environment translates its state from $s_t$ to $s_{t+1}$ responding to the action and the agent also obtains a reward $r_t\in R$. 
The agent's objective is to learn an optimal policy  $\pi^*$ so as to maximize the expected accumulative rewards. 
Let $\eta(\pi)$ denote the expected cumulative reward:
\begin{eqnarray}
\eta(\pi)=\mathbb{E}_{s_0,a_0,\cdots}\bigg[\sum_{t=0}^{T}\gamma^tr(s_t, a_t)\bigg]
\end{eqnarray} 
where $\gamma\in[0,1]$ is a discounting factor.
The deterministic policy is defined as:
\begin{eqnarray}
\pi: S \rightarrow A
\end{eqnarray}
which is a mapping from the state space to the action space. 
Let $\pi(a_t|s_t)$ represent the the conditional probability that the agent takes action $a_t$ given the states $s_t$.

To apply the above RL framework to ZO optimization,  we first need define some features to represent the state of ZO gradient, a set of actions to represent the sampling rule, and a reward function. Then, we should choose an appropriate algorithm to find the optimal policy $\pi^*$ to maximize our expected cumulative reward. In the next section, we will discuss these in detail.

\section{Learning Sampling Policy in Zeroth-Order Optimization}
Considering that the learned sampling policy may perform better in ZO optimization compared to random sampling, we learn the sampling policy  $\pi$  in ZO optimization. 
We observe that the execution of ZO optimization algorithm can be viewed as the execution of a fixed policy in a MDP: the state consists of the current function value and ZO gradients evaluated at the current and past function values, the action is the step vector that is used to update the current parameter, and the transition probability is partially characterized by the parameter update formulation.
The policy that is executed corresponds precisely to the choice of $\pi$ used by the ZO optimization algorithm.
Thus, we use the RL to learn the policy $\pi$.
For this purpose, we need to define the reward function that should reward those policies that exhibit good behavior during execution.
Since the performance metric of interest to the ZO optimization algorithm is the speed of convergence, the reward function should reward policies that converge quickly.
To this end, assuming the goal is to minimize the objective function, we define the reward in a given state as the value of the ZO gradient. This will encourage the policy to reach the minimum of the objective function as soon as possible.
Therefore, the learned sampling policy by maximizing expected cumulative reward can effectively reduce the variances of ZO gradient.

In the following, we first introduce the principle of our ZO-RL algorithm. Then, we introduce the network structure and the batch normalization technique which are  used in our  ZO-RL algorithm.

\subsection{Principle of our ZO-RL Algorithm}
Since the action space is continuous in ZO optimization, we use the deterministic policy.
Compared with the stochastic policy gradient algorithm, the deterministic policy has the advantages of requiring less data to be sampled and high efficiency of the algorithm, and performs stably in a series of tasks with continuous action space.
Thus, to find the optimal policy, we use deep deterministic policy gradient (DDPG) to learn query policy $\pi$.
DDPG is an actor-critic and model-free algorithm for RL over continuous action spaces and output deterministic actions in a stochastic environment to maximize cumulative rewards.

The DDPG has two neural network function approximators. 
One is called the actor network $\pi(s|\theta^{\pi})$ with weights $\theta^{\pi}$ which learns a policy function of mapping a state to a deterministic action. 
The other is called the critic network $Q(s,a|\theta^{Q})$ with weights $\theta^{Q}$, which can learn a state-action value function and its input consists of a state and action. 
The actor-critic methods combine the advantages of critic-based and actor-based methods. These methods estimate the parameters of the critic and the critic simultaneously. The critic learns a value function, which is used to measure whether the current action is improved compared with the policy’s default behavior.
The parameters of the actor’s policy are updated in a direction advised by the critic evaluation. The parameters of two structures are simultaneously estimated to find out the optimal policy.
In addition, DDPG creates a copy of the actor and critic networks, $Q'(s,a|\theta^{Q'})$ and $\pi'(s|\theta^{\pi'})$ respectively, that are used for calculating the target values. The weights of these target networks are then updated by making them slowly track the learned networks: $\theta'\rightarrow \tau\theta'+(1-\tau)\theta'$ with $\tau\ll 1$. This means that the target values are constrained to change slowly, greatly improving the stability of learning. This simple change moves the relatively unstable problem of learning the action-value function closer to the case of supervised learning.
The updates at each iteration contain the critic update and actor update. 
Updating the critic is to minimize a squared-error loss $L$:
\begin{eqnarray}
\min_{\theta^Q}L=\frac{1}{N} \sum_{i=1}^{N}(y_i-Q(s_i,a_i|\theta^Q))^2
\end{eqnarray}
where $N$ is the mini batch size, $Q(\cdot|\theta^Q)$ represents a parameterized state-action value function, and $y_i$ represents the TD target denoted as 
\begin{eqnarray}
y_i=r_i+\gamma Q'(s_{i+1},\pi'(s_{i+1}|\theta^{\pi'})|\theta^{Q'})
\end{eqnarray}
Updating the actor can help maximize the cumulative reward using a sampled policy gradient:
\begin{eqnarray}
\nabla_{\theta^{\pi}}J=\frac{1}{N}\sum_{i=1}^{N}\nabla_{a}Q(s,a|\theta^Q)|_{s=s_i,a=\pi(s_i)}\nabla_{\theta^{\pi}}\pi(s|\theta^\pi)|_{s_i}
\end{eqnarray}
where $J$ and $\pi(\cdot|\theta^{\pi})$ represent the expected cumulative reward and parameterized policy function respectively.
We summarize our ZO-RL optimization algorithm in Algorithm \ref{alg:2}.

\begin{algorithm}[H]
	\caption{Reinforcement Learning Based Zeroth-Order (ZO-RL) Algorithm} 	\label{alg:2}
	\renewcommand{\algorithmicrequire}{\textbf{Input:}}
	\renewcommand{\algorithmicensure}{\textbf{Output:}}
	\begin{algorithmic}[1]
		\REQUIRE Randomly initialize critic network $Q(s,a|\theta^{Q})$ and actor $\pi(s|\theta^{\pi})$. Hyper-parameter $\mu$, $q$ and $\eta$.
		\ENSURE $x\in\mathbb{R}^d$ 
		\STATE Using DDPG to learn optimal sampling policy $\pi^*$ from the standard Gaussian distribution.
		\FOR {$k=0$ to $K-1$} 
		\STATE Select action $a_k=\pi^*(s_k|\theta^{\pi^*})$ according to the sampling policy $\pi^*$. 
		\STATE Calculating the ZO gradient estimator $\hat{\nabla} f(x_k)$.
		\STATE Obtain the next update $x_{k+1}=x_k-\eta\hat{\nabla} f(x_k)$.
		\ENDFOR
	\end{algorithmic}
\end{algorithm}

\subsection{Network Structure}
The choice of the structure of the critic and actor nets is important because they are not only function approximators but also part of the feature learning.
We choose the convolutional neural network (CNN) \cite{sezer2018algorithmic} both for the critic net and the actor net. 
CNN has a large variety of applications in image-classification, video-recognition, and also ``natural language processing". 
Generally, CNN consists of three types of layers: the convolutional, pooling and multilayer perceptron (MLP).
A convolutional and pooling layer are structurally successive. In the ``convolution layer", the convolution operation is carried out, results are passed to the pooling layer. In the pooling layer, both the “number of parameters" and ``the spatial size of representation" are reduced. In the last convolutional or pooling layer, the data forms a  one-dimensional vector and is connected to an MLP. In other word, convolutional and pooling layers perform an implicit feature extraction, and MLP performs a traditional classifier. 
The input matrix can be viewed as a  one-dimensional image with $n$ channels, that is, each technical indicator represents a channel. 
For the critic network, the input is a state $s_t$ and action $a_t$, and its output is state-action value function $Q(s_i,a_i|\theta^Q)$.
For the actor network, the input is the state $s_t$ and the output is the probabilities of taking actions.

\subsection{Batch Normalization}
The parameters of the optimizer have different descent rates in different dimensions and the range may be different in different environments.
This may make it difficult for the network to learn efficiently and find hyper-parameters that generalize the scale of state values in different environments.
One way to address this issue is to manually scale features so that they are in a similar range across environments and units. We address this problem by adapting one of the latest techniques in deep learning, called batch normalization \cite{santurkar2018does}.
This technique normalizes each dimension of a sample in a mini-batch to have unit mean and variance. In addition, batch normalization maintains a running average of the mean and variance to be used for normalization during testing. In deep networks, batch normalization is used to minimize the covariance bias during training, by ensuring that each layer receives whitened inputs.

\section{Combining Our ZO-RL with Existing ZO Optimization Algorithms}
In this section, we discuss how to combine our ZO-RL with an existing ZO algorithm based on the parameter update rule or the  learned  sampling  distribution.
For example, \cite{ruan2019learning} proposed a ZO optimization algorithm called ZO-LSTM, which  replaces parameter update rule as well as guided sampling rule to sample the perturbed samples with learned recurrent neural networks (RNN).
Especially, they updated the parameter through a Long Short-Term Memory (LSTM) network called UpdateRNN:
\begin{eqnarray}
x_k=x_{k-1}+\textrm{UpdateRNN}(\hat{\nabla} f(x_k))
\end{eqnarray}
where $x_k$ is the optimizer parameter at iteration $k$.
UpdateRNN can reduce the negative impact of high variances of ZO gradient estimator due to long-term dependence, in addition to learning to compute parameter updates adaptively by exploring the loss landscape.
They use another LSTM network called QueryRNN to learn the sampling rules for distributions.
They dynamically predict the converiance matrix $\Sigma_k$:
\begin{eqnarray}
\Sigma_k=\textrm{QueryRNN}([\hat{\nabla} f(x_k),\Delta x_{k-1}])
\end{eqnarray}
QueryRNN can increase the sampling probability in the direction of the bias of the estimated gradient or the parameter update of the previous iteration.

Although they considered both the sampling distribution and the parameter update rule, they still used random sampling for the perturbed samples.
Using the learned sampling policy on the learned sampling distribution can further speed up the convergence of ZO optimization.
Thus, we can combine our ZO-RL algorithm with ZO-LSTM algorithm.
Especially, we first train the UpdateRNN using standard Gaussian random vectors as query directions.
Then we freeze the parameters of the UpdateRNN and train the QueryRNN.
Finally, we use the previous work as a warm start and use our ZO-RL in the pre-learning distribution to learn the sampling policy.
In addition, other ZO algorithms based on parameter update rules such as \cite{lian2016comprehensive,chen2017zoo}, can be directly combined with our ZO-RL algorithm.

\begin{figure*}[!t]
	\centering	
	\centering		
	\begin{subfigure}[b]{0.23\textwidth}
		\centering
		\includegraphics[width = 1.15\textwidth]{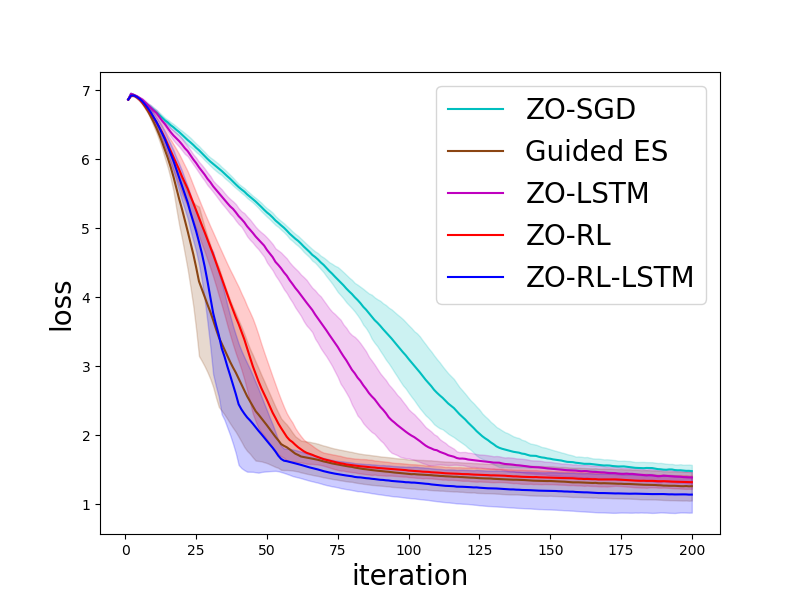}
		\caption{MNIST TEST 1}
	\end{subfigure}
	\begin{subfigure}[b]{0.23\textwidth}
		\centering
		\includegraphics[width = 1.15\textwidth]{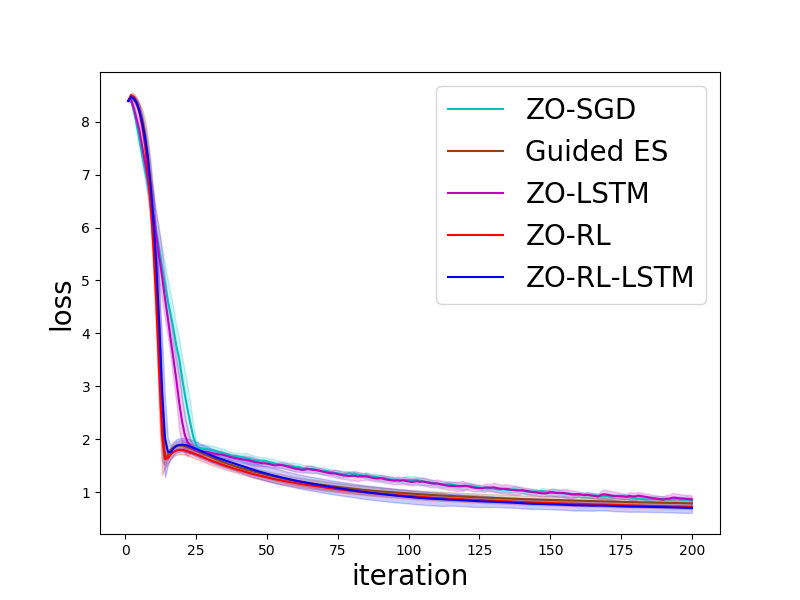}
		\caption{MNIST TEST 2}
	\end{subfigure}
	\begin{subfigure}[b]{0.23\textwidth}
		\centering
		\includegraphics[width = 1.15\textwidth]{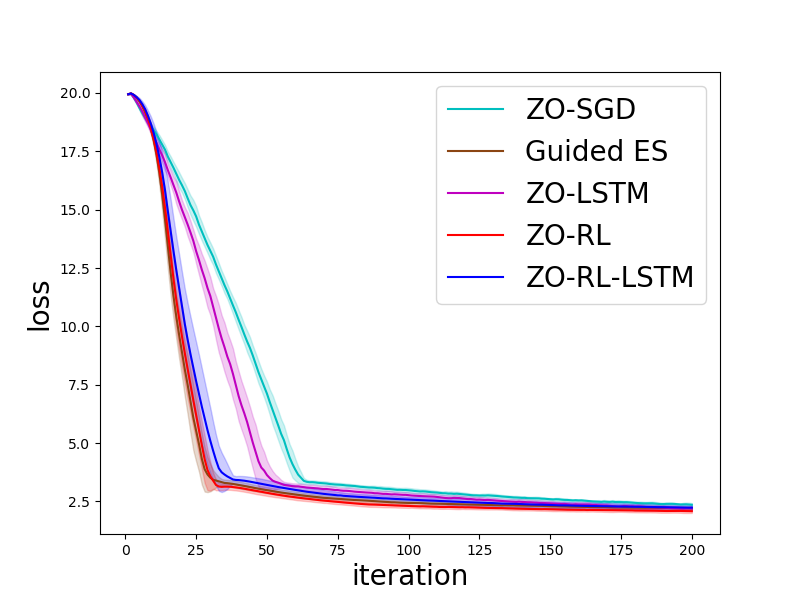}
		\caption{MNIST TEST 3}
	\end{subfigure}
	\begin{subfigure}[b]{0.23\textwidth}
		\centering
		\includegraphics[width = 1.15\textwidth]{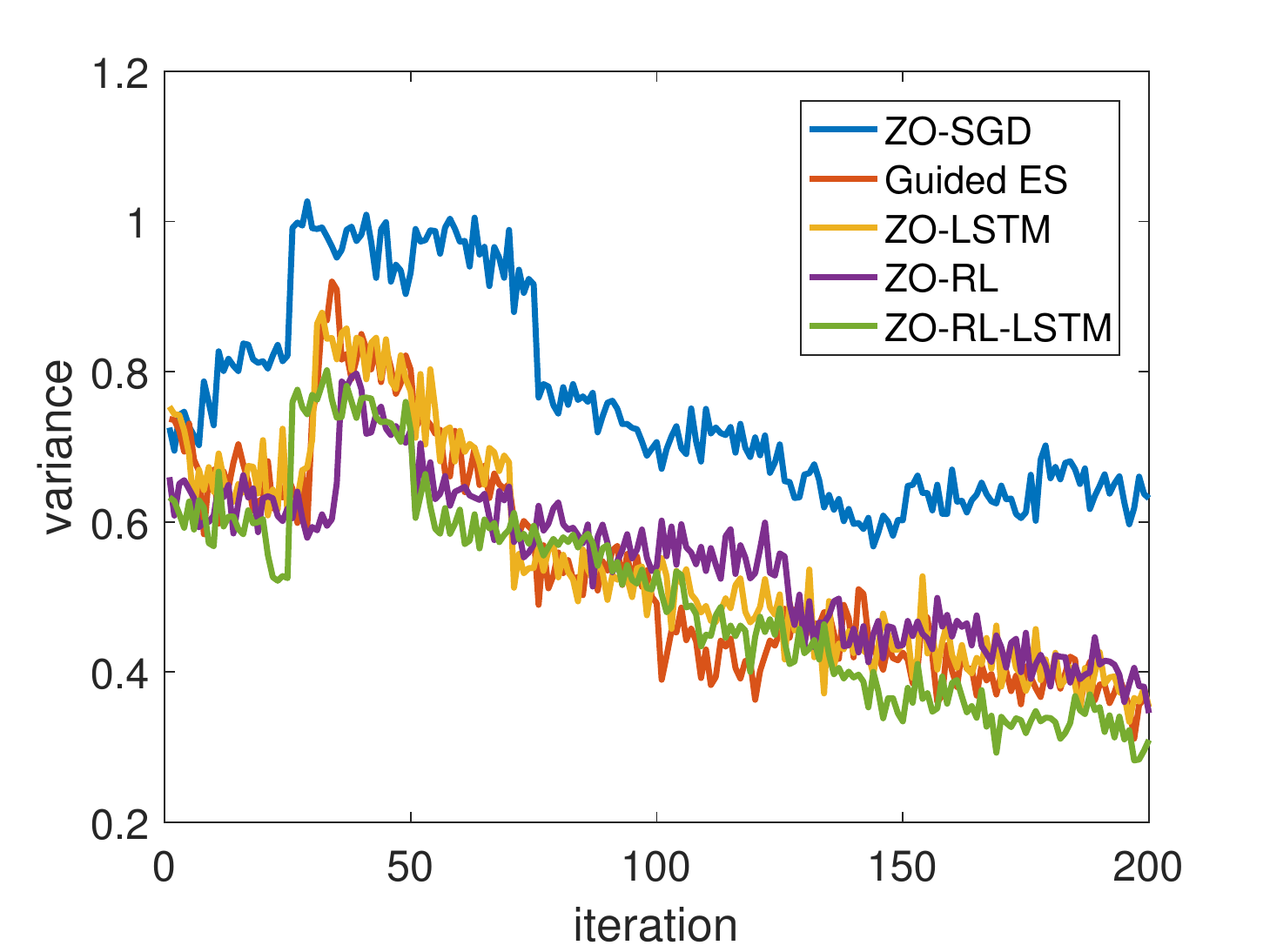}
		\caption{MNIST TEST 3}
	\end{subfigure}	
	\caption{Adversarial attack to black-box models in the SGD setting.}
	\label{fig:2}
\end{figure*}

\begin{figure*}[!t]
	\centering		
	\begin{subfigure}[b]{0.23\textwidth}
		\centering
		\includegraphics[width = 1.15\textwidth]{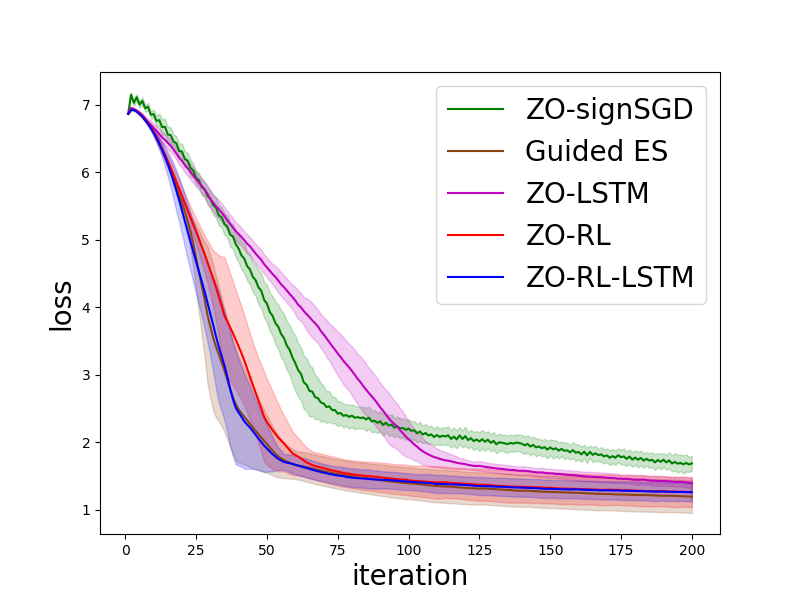}
		\caption{MNIST TEST 1}
	\end{subfigure}
	\begin{subfigure}[b]{0.23\textwidth}
		\centering
		\includegraphics[width = 1.15\textwidth]{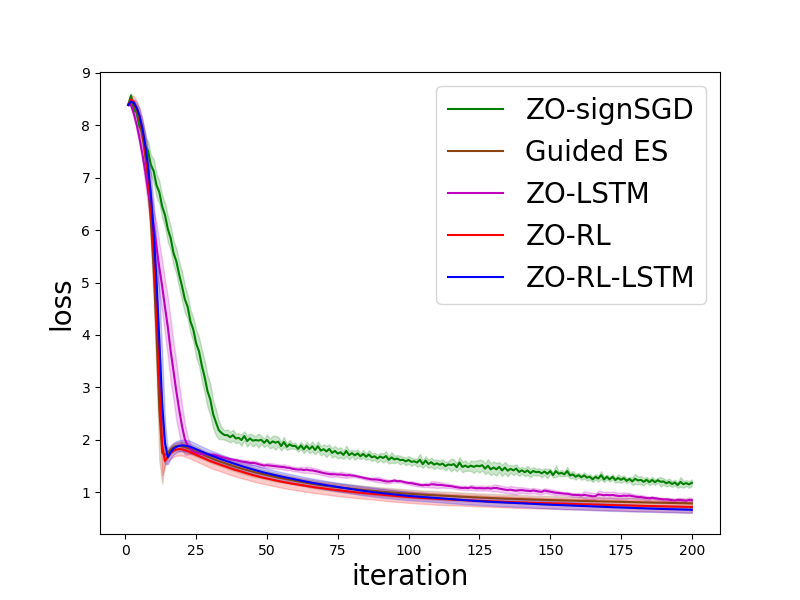}
		\caption{MNIST TEST 2}
	\end{subfigure}
	\begin{subfigure}[b]{0.23\textwidth}
		\centering
		\includegraphics[width = 1.15\textwidth]{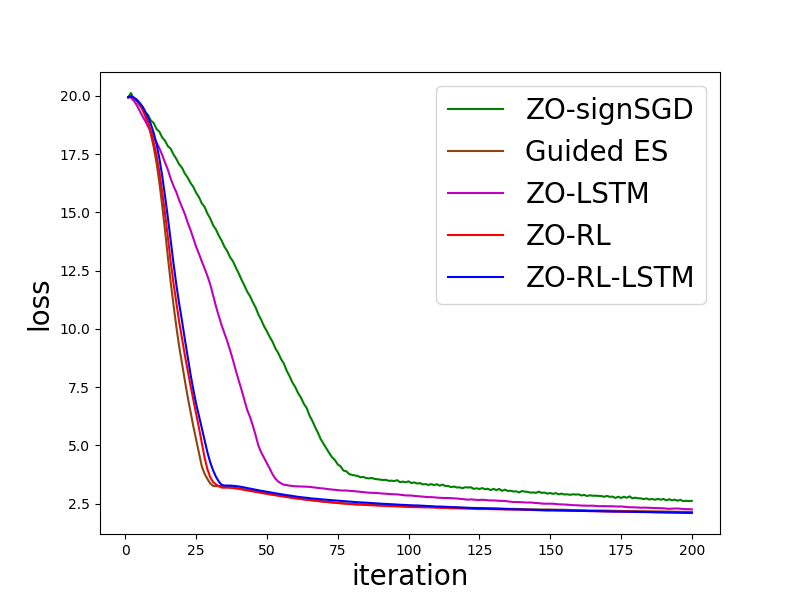}
		\caption{MNIST TEST 3}
	\end{subfigure}
	\begin{subfigure}[b]{0.23\textwidth}
		\centering
		\includegraphics[width = 1.15\textwidth]{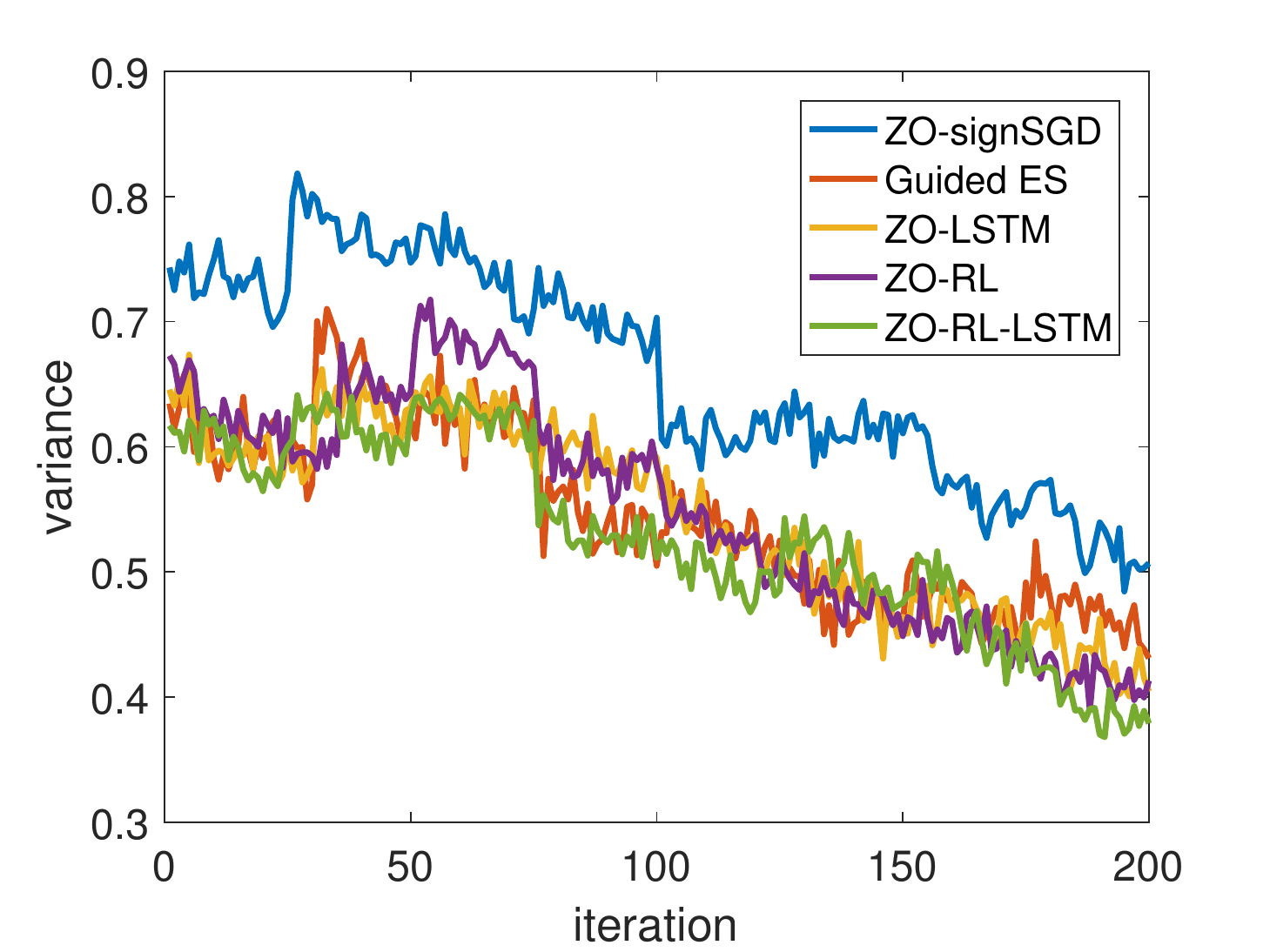}
		\caption{MNIST TEST 3}
	\end{subfigure}		
	\caption{Adversarial attack to black-box models in the signSGD setting.}
	\label{fig:3}
\end{figure*}

\begin{figure*}[!t]
	\centering		
	\begin{subfigure}[b]{0.23\textwidth}
		\centering
		\includegraphics[width = 1.15\textwidth]{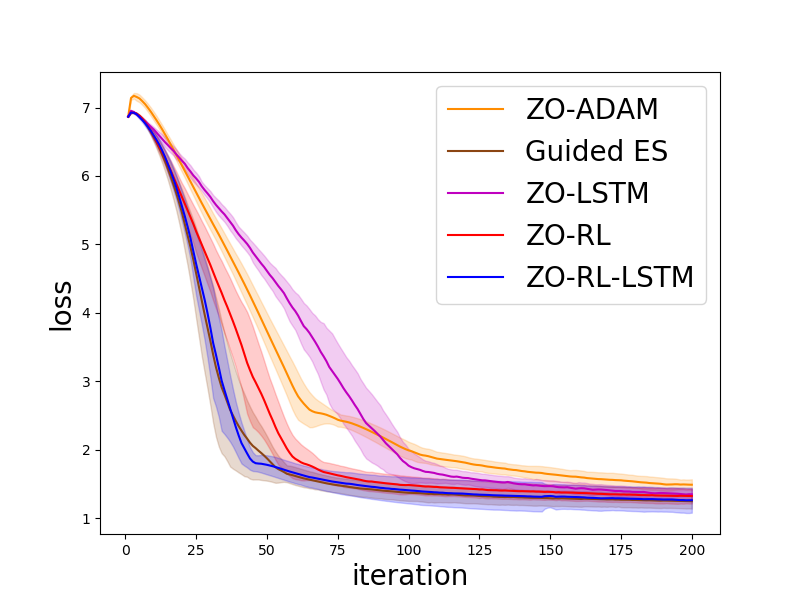}
		\caption{MNIST TEST 1}
	\end{subfigure}
	\begin{subfigure}[b]{0.23\textwidth}
		\centering
		\includegraphics[width = 1.15\textwidth]{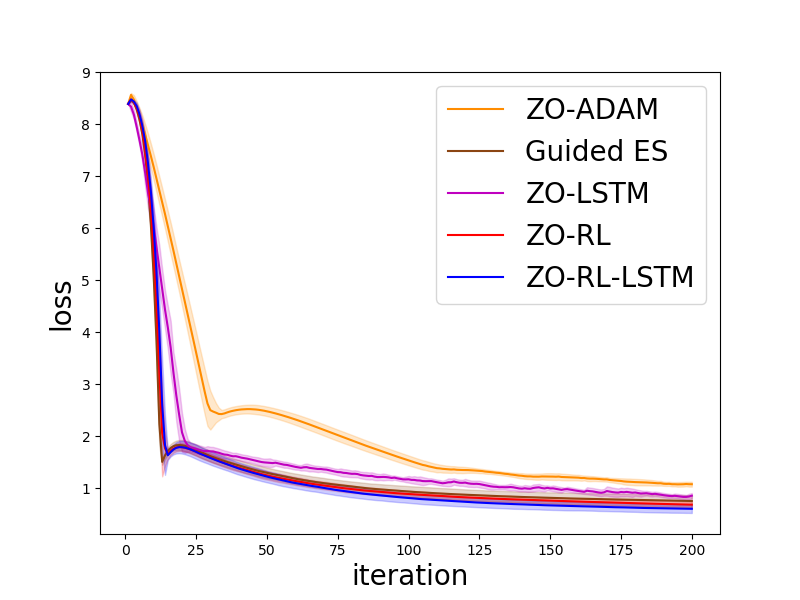}
		\caption{MNIST TEST 2}
	\end{subfigure}
	\begin{subfigure}[b]{0.23\textwidth}
		\centering
		\includegraphics[width = 1.15\textwidth]{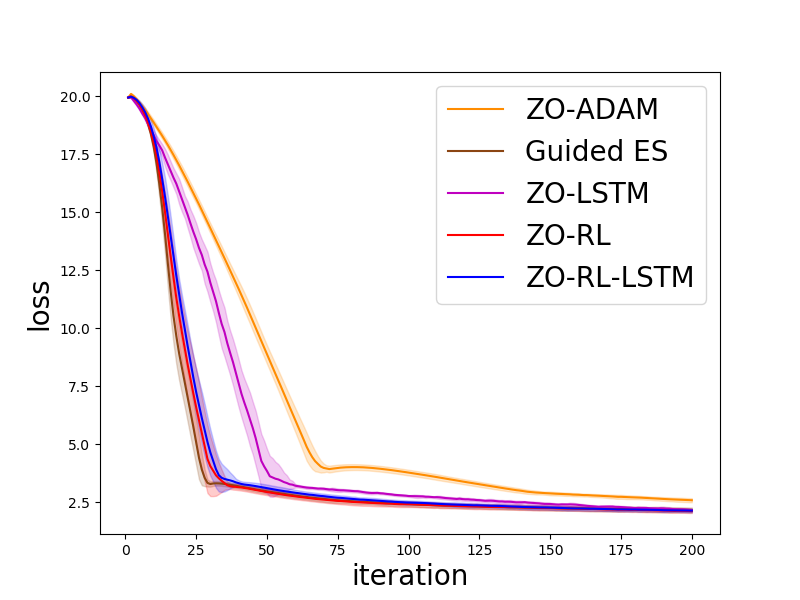}
		\caption{MNIST TEST 3}
	\end{subfigure}
	\begin{subfigure}[b]{0.23\textwidth}
		\centering
		\includegraphics[width = 1.15\textwidth]{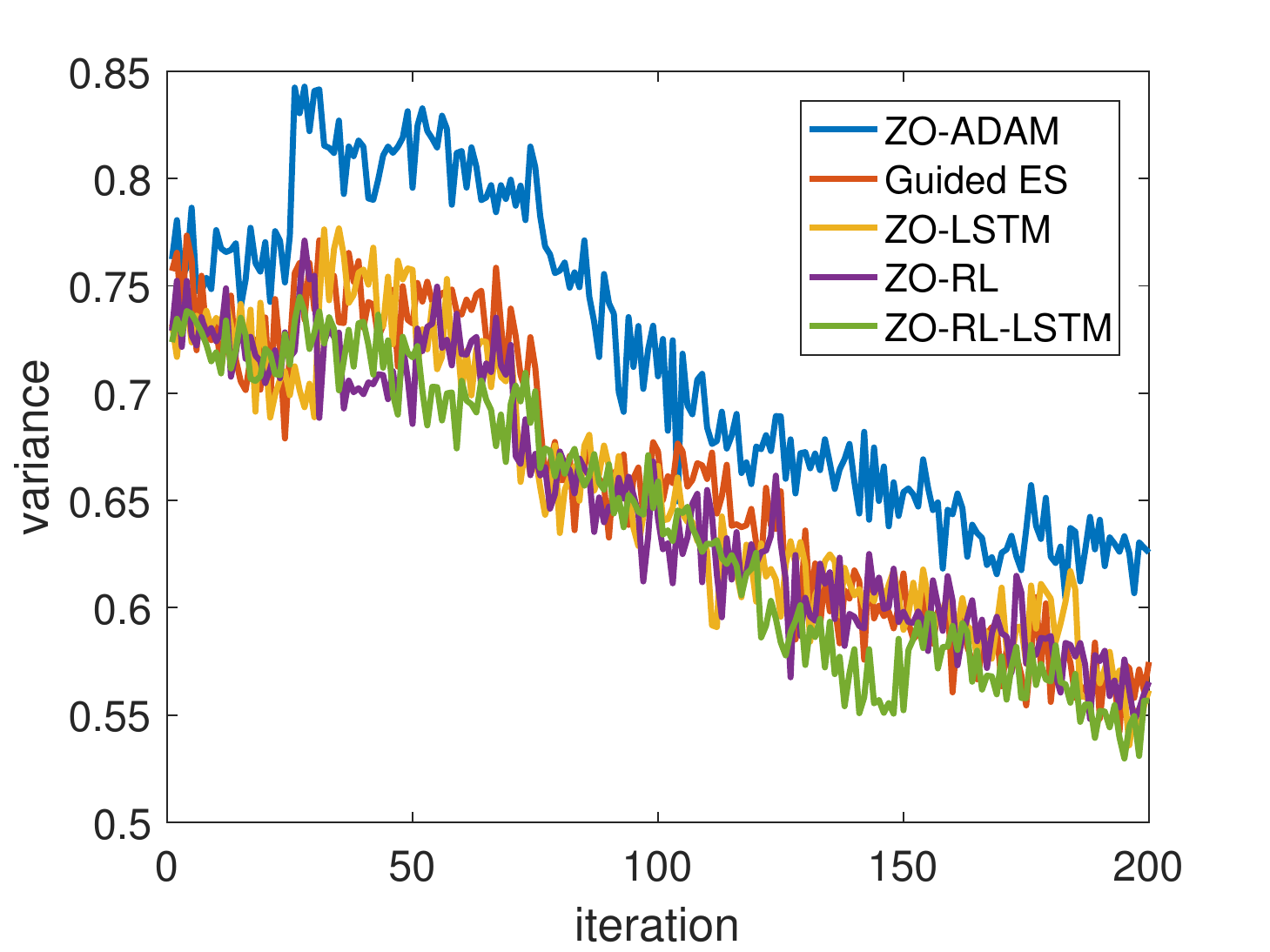}
		\caption{MNIST TEST 3}
	\end{subfigure}	
	\caption{Adversarial attack to black-box models in the ADAM setting.}
	\label{fig:4}
\end{figure*}

\section{Experiments}
In this section, we empirically demonstrate the superiority of our proposed ZO optimizer on a practical application (black-box adversarial attack on MNIST dataset) and a synthetic problem (non-convex optimization problems on benchmark datasets).
To show the effectiveness of the learned sampling policy, we compare our ZO optimizer with existing ZO optimization algorithms.
Specifically, we consider the convergence behavior and change in variances of ZO optimizer under three different parameter update settings.
The three different parameter update settings are summarized as follows:
\begin{enumerate}
	\item SGD \cite{ghadimi2013stochastic}: A gradient descent algorithm based on the ZO gradient estimators.
	\item signSGD \cite{liu2018signsgd}: A gradient descent algorithm based on the sign of the ZO gradient estimators.
	\item ADAM \cite{chen2017zoo}: A gradient descent algorithm based on adaptive estimates of lower-order moments.
\end{enumerate}

We obtain ZO gradient estimator along sampled directions via ZO Oracle.
Since our algorithm is the first one to learning the sampling policy, we compare the performance of the ZO gradient estimators sampled form the standard Gaussian distribution and two learned Gaussian distributions, \textit{i.e.} using different covariance matrix $\Sigma$.
In addition, we compare the algorithm of synchronous learning sampling and distribution policy by combining our algorithm with other algorithms.
The five algorithms for calculating ZO gradient estimators are summarized as follows:
\begin{enumerate}
	\item ZO-GS \cite{wang2019zeroth}: Randomly Sampling the perturbed vectors ${u_i}$ from a standard Gaussian distribution.
	\item ZO-LSTM \cite{ruan2019learning}: They learned the Gaussian sampling rule and dynamically predicted the covariance matrix $\Sigma$ for query directions with recurrent neural networks.
	\item Guided ES \cite{maheswaranathan2019guided}: They let the covariance matrix  $\Sigma$ be related with the recent history of surrogate gradients during optimization.
	\item ZO-RL: Our proposed ZO algorithm learns the sampling policy through reinforcement learning.
	\item ZO-RL-LSTM: Our proposed ZO algorithm combined with ZO-LSTM to learn sampling policy on a learned Gaussian distribution.
\end{enumerate}

\subsection{Implementation}
For each task, we tune the hyper-parameters of baseline algorithms to report the best performance.
We coarsely tune the constant $\delta$ on a logarithmic range $\{0.01; 0.1; 1; 10; 100; 1000\}$ and set the learning rate of baseline algorithms to $\eta=\delta/d$, where $d$ is the dimension of dataset.
For ADAM, we tune $\beta_1$ values over $\{0.9, 0.99\}$ and $\beta_2$ values over $\{0.99, 0.996, 0.999\}$.
We set the smoothing parameter $\mu=0.01$ in all experiments.
To ensure fair comparison, all optimizers are using the same number of query directions to obtain ZO gradient estimator at each iteration.

\subsection{Adversarial Attack to Black-box Models}
We consider generating adversarial examples to attack black-box DNN image classifier and formulate it as a zeroth-order optimization problem.
The targeted DNN image classifier $F(x)=[F_1,F_2,\cdots,F_{K}]$ takes as input an image $x\in[0,1]^d$ and outputs the prediction scores of $K$ classes.
Given an image $x_0\in[0,1]^d$ and its corresponding true label $t_0\in[1,2,\cdots,K]$,
an adversarial example $x$  is visually similar to the original image $x_0$ but leads the targeted model $F$ to make wrong prediction other than $t_0$.
The black-box attack loss is defined as:
\begin{eqnarray}
\min_{x} \max \{F_{t_0}(x)-\max_{j\neq t_0}F_j(x),0\}+c\Vert x-x_0\Vert_{p}
\end{eqnarray}
The first term is the attack loss which measures how successful the adversarial attack is and penalizes correct prediction by the targeted model. The second term is the distortion loss ($p$-norm of added perturbation) which enforces the perturbation added to be small and $c$ is the regularization coefficient.
In our experiment, we use $\ell_1$ norm (i.e., $p = 1$), and set $c = 0.1$ for MNIST attack task.
Due to the black-box setting, one can only compute the function values of the above objective, which leads to ZO optimization problems \cite{chen2017zoo}.
Note that attacking each sample $x_0$ in the dataset corresponds to a particular ZO optimization problem instance, which motivates us to train a ZO optimizer offline with a small subset, and apply it to online attack to other samples with faster convergence (which means lower query complexity) and lower final loss (which means less distortion).
We randomly select 50 images that are correctly classified by the targeted model in each test set to train the optimizer and select another 50 images to test the learned optimizer.
The number of sampled query directions is set to $q = 20$ for MNIST, and the optimizer is allowed to run 200 steps.

\subsection{Non-Convex Optimization Problems}
We consider a binary classification problem with a non-convex least squared loss function $\min_{w\in\mathbb{R}^d}  \frac{1}{n}\sum_{i=1}^{n}(y_i-1/(1+e^{-w^Tx_i}))^2$.
Here $(x_i,y_i)$ is the $i$th data sample containing feature $x_i\in\mathbb{R}^d$ and label $y_i\in\{ -1,1 \}$.
We compare the algorithms on benchmark datasets (heat scale, german and a9a\footnote{\url{http://archive.ics.uci.edu/ml/datasets.html}}).
All the algorithms can only access to the oracle of function value evaluations.
We use the same set of hyper-parameters for different datasets and repeated runs in the experiments.
The number of query directions are set to $q=20$.
For each dataset, we repeat the experiment 10 times and report the average and the standard deviation.
At each iteration of training, the optimizer is allowed to run 200 steps.

\begin{figure*}[!t]
	\centering	
	\begin{subfigure}[b]{0.23\textwidth}
		\centering
		\includegraphics[width = 1.15\textwidth]{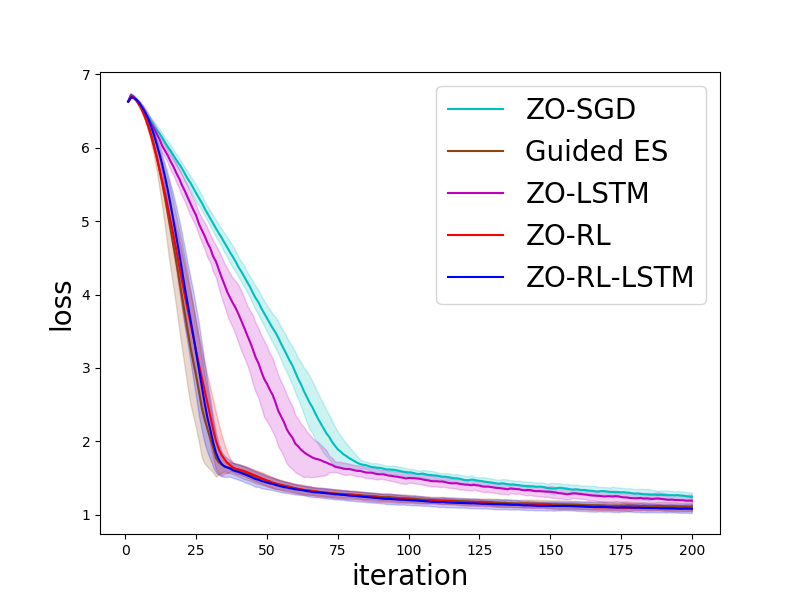}
		\caption{heat scale}
	\end{subfigure}
	\begin{subfigure}[b]{0.23\textwidth}
		\centering
		\includegraphics[width = 1.15\textwidth]{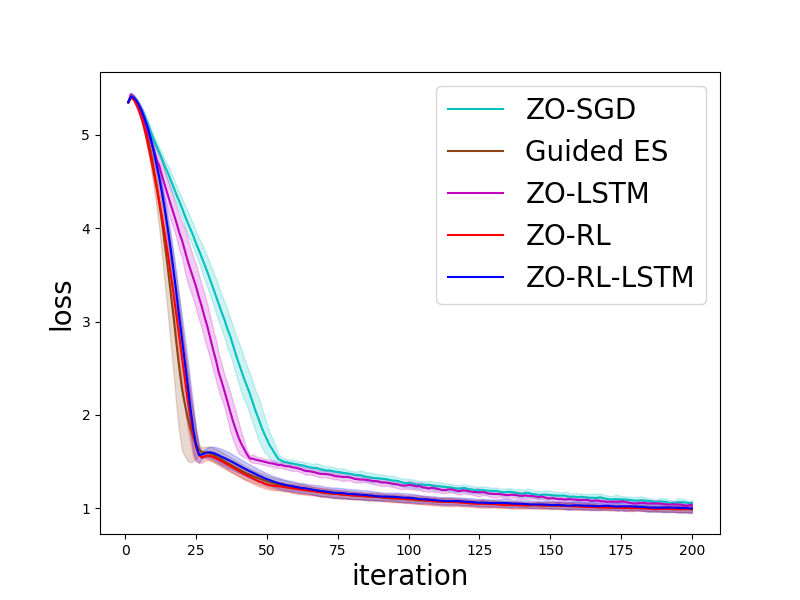}
		\caption{german}
	\end{subfigure}
	\begin{subfigure}[b]{0.23\textwidth}
		\centering
		\includegraphics[width = 1.15\textwidth]{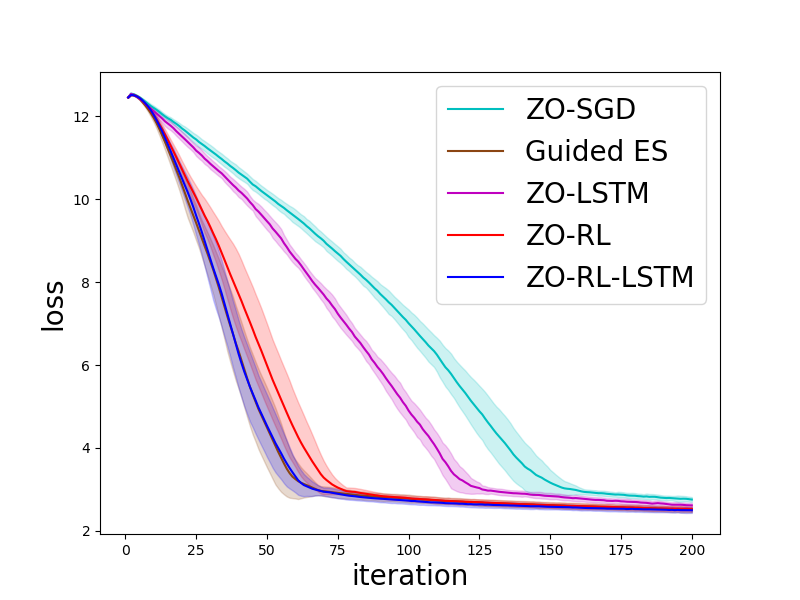}
		\caption{a9a}
	\end{subfigure}
	\begin{subfigure}[b]{0.23\textwidth}
		\centering
		\includegraphics[width = 1.15\textwidth]{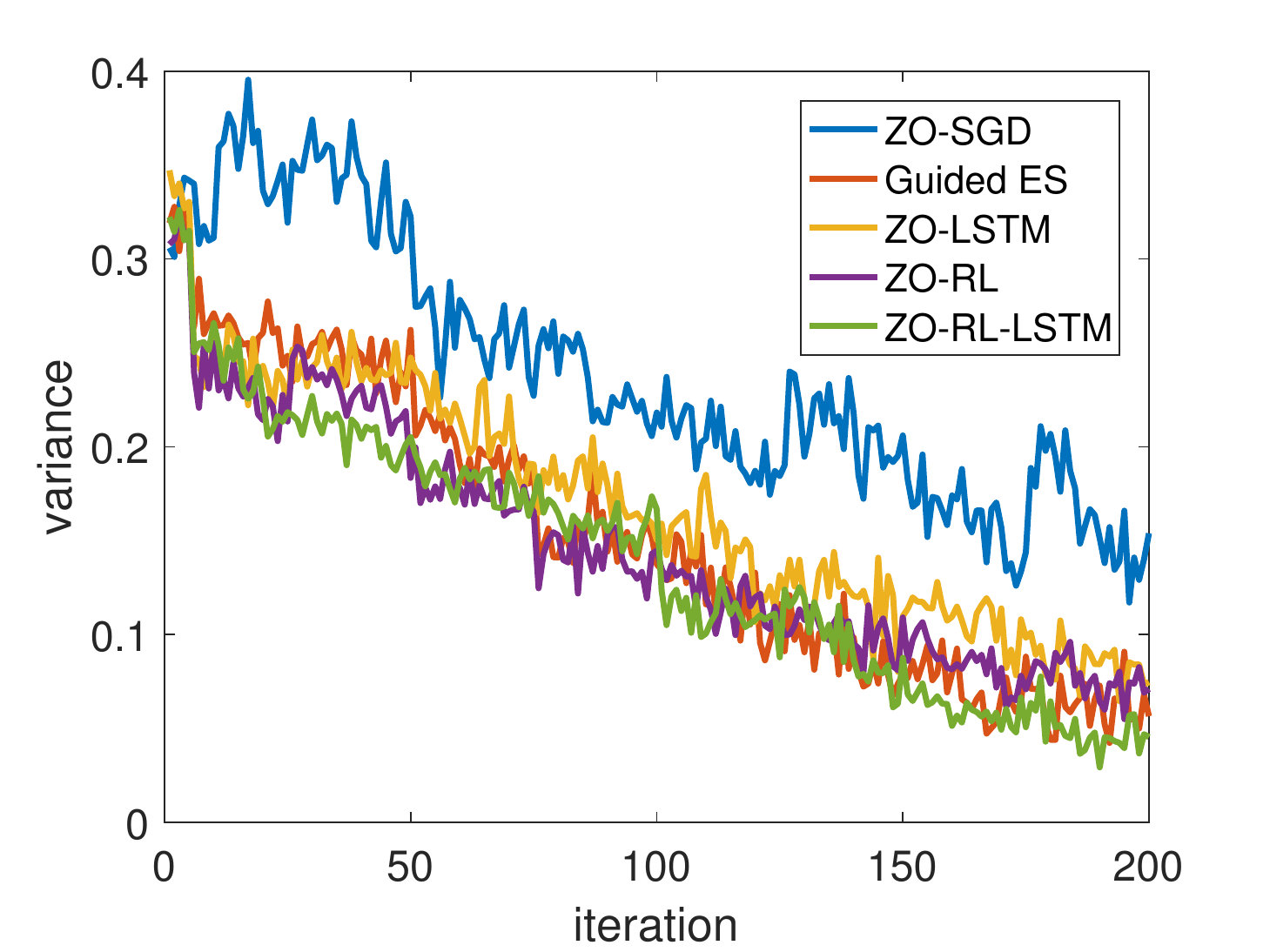}
		\caption{a9a}
	\end{subfigure}
	\caption{Non-convex optimization problems in the SGD setting.}
	\label{fig:5}
\end{figure*}

\begin{figure*}[!t]
	\centering	
	\begin{subfigure}[b]{0.23\textwidth}
		\centering
		\includegraphics[width = 1.15\textwidth]{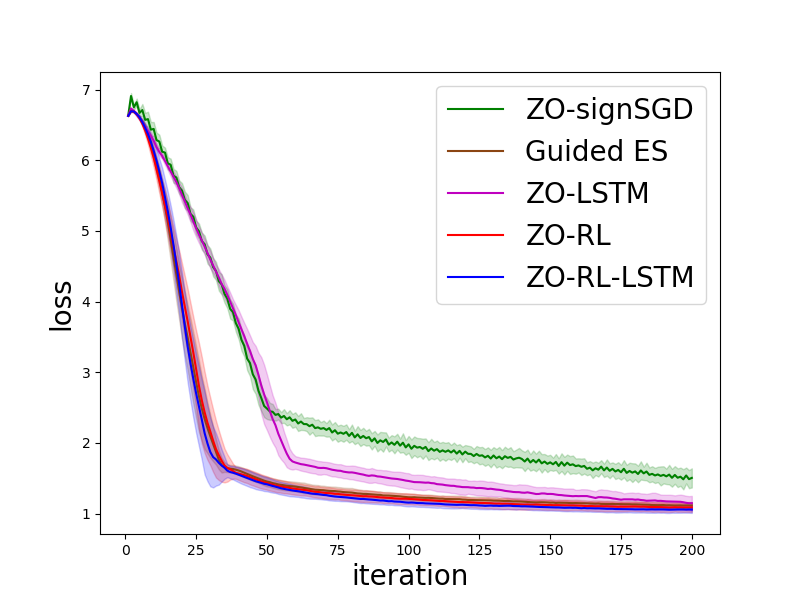}
		\caption{heat scale}
	\end{subfigure}
	\begin{subfigure}[b]{0.23\textwidth}
		\centering
		\includegraphics[width = 1.15\textwidth]{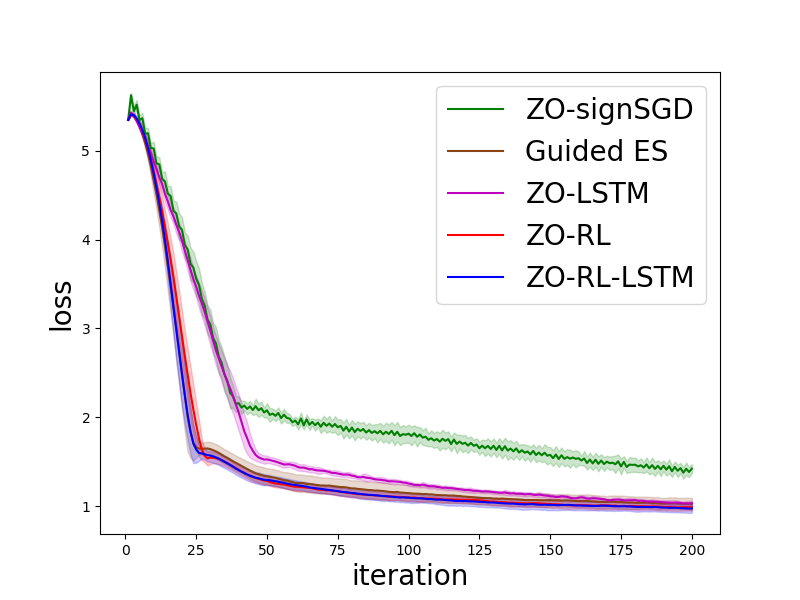}
		\caption{german}
	\end{subfigure}
	\begin{subfigure}[b]{0.23\textwidth}
		\centering
		\includegraphics[width = 1.15\textwidth]{eps/signSGD/signSGD_2_2.png}
		\caption{a9a}
	\end{subfigure}
	\begin{subfigure}[b]{0.23\textwidth}
		\centering
		\includegraphics[width = 1.15\textwidth]{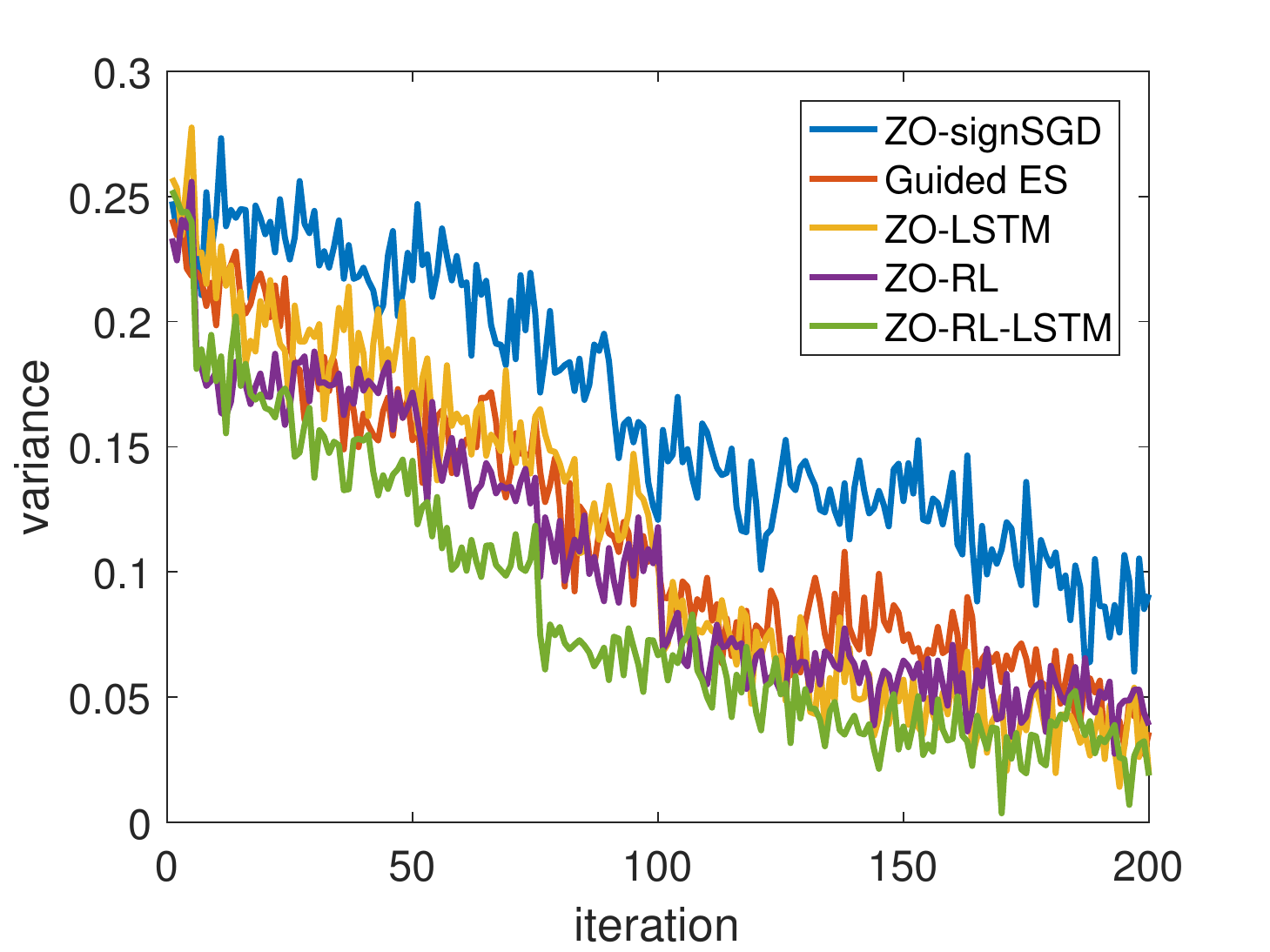}
		\caption{a9a}
	\end{subfigure}
	\caption{Non-convex optimization problems  in the signSGD setting.}
	\label{fig:6}
\end{figure*}

\begin{figure*}[!t]
	\centering	
	\begin{subfigure}[b]{0.23\textwidth}
		\centering
		\includegraphics[width = 1.15\textwidth]{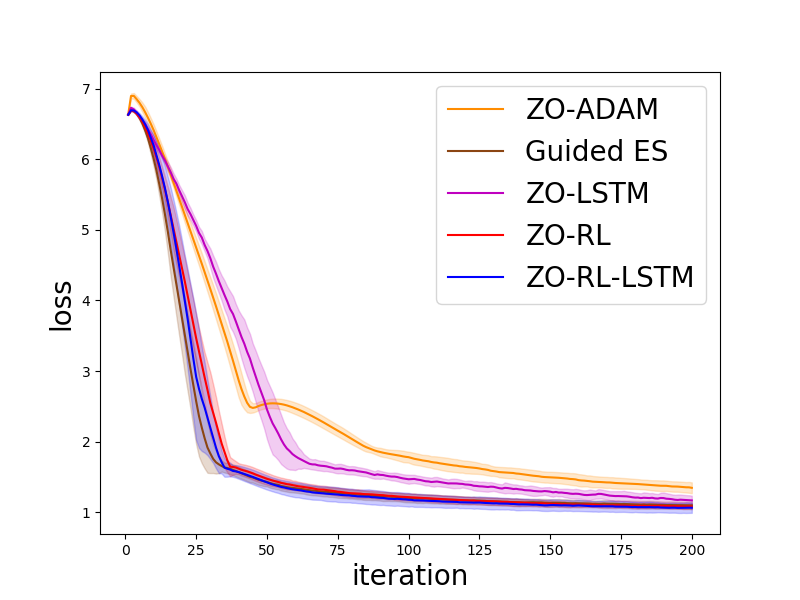}
		\caption{heat scale}
	\end{subfigure}
	\begin{subfigure}[b]{0.23\textwidth}
		\centering
		\includegraphics[width = 1.15\textwidth]{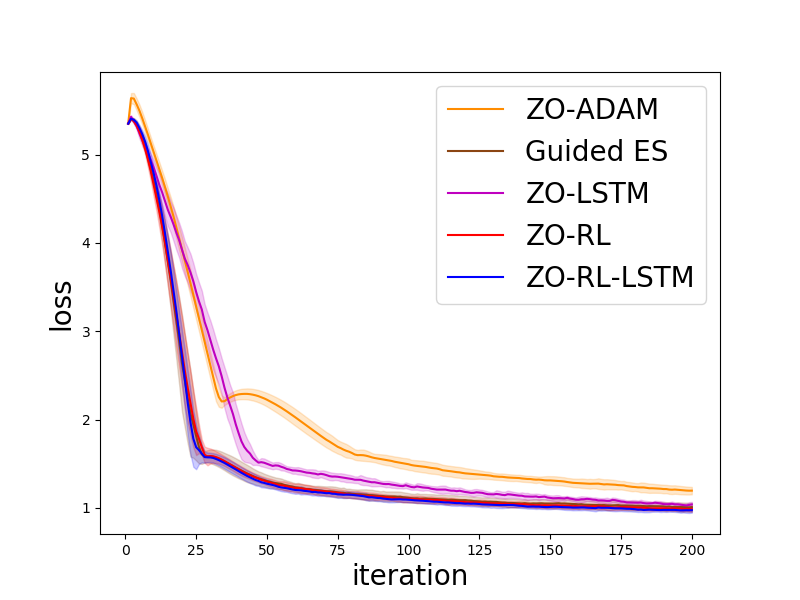}
		\caption{german}
	\end{subfigure}
	\begin{subfigure}[b]{0.23\textwidth}
		\centering
		\includegraphics[width = 1.15\textwidth]{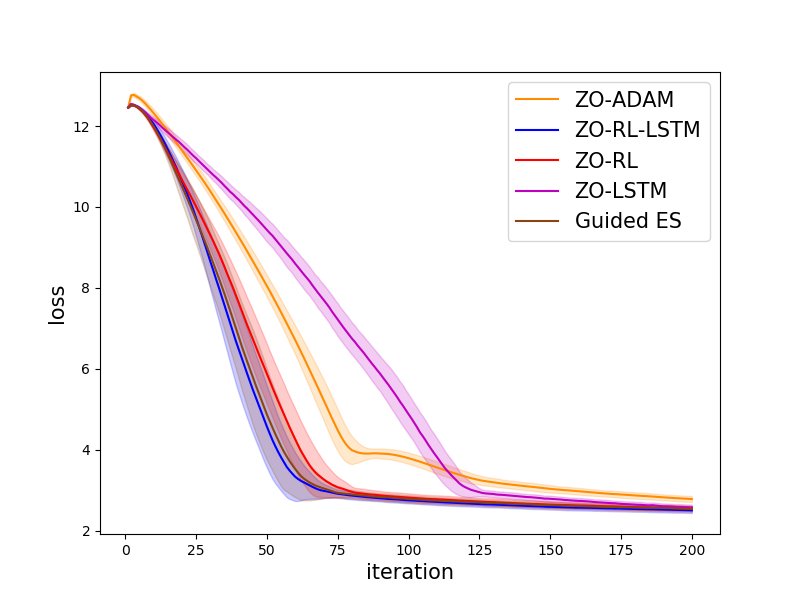}
		\caption{a9a}
	\end{subfigure}
	\begin{subfigure}[b]{0.23\textwidth}
		\centering
		\includegraphics[width = 1.15\textwidth]{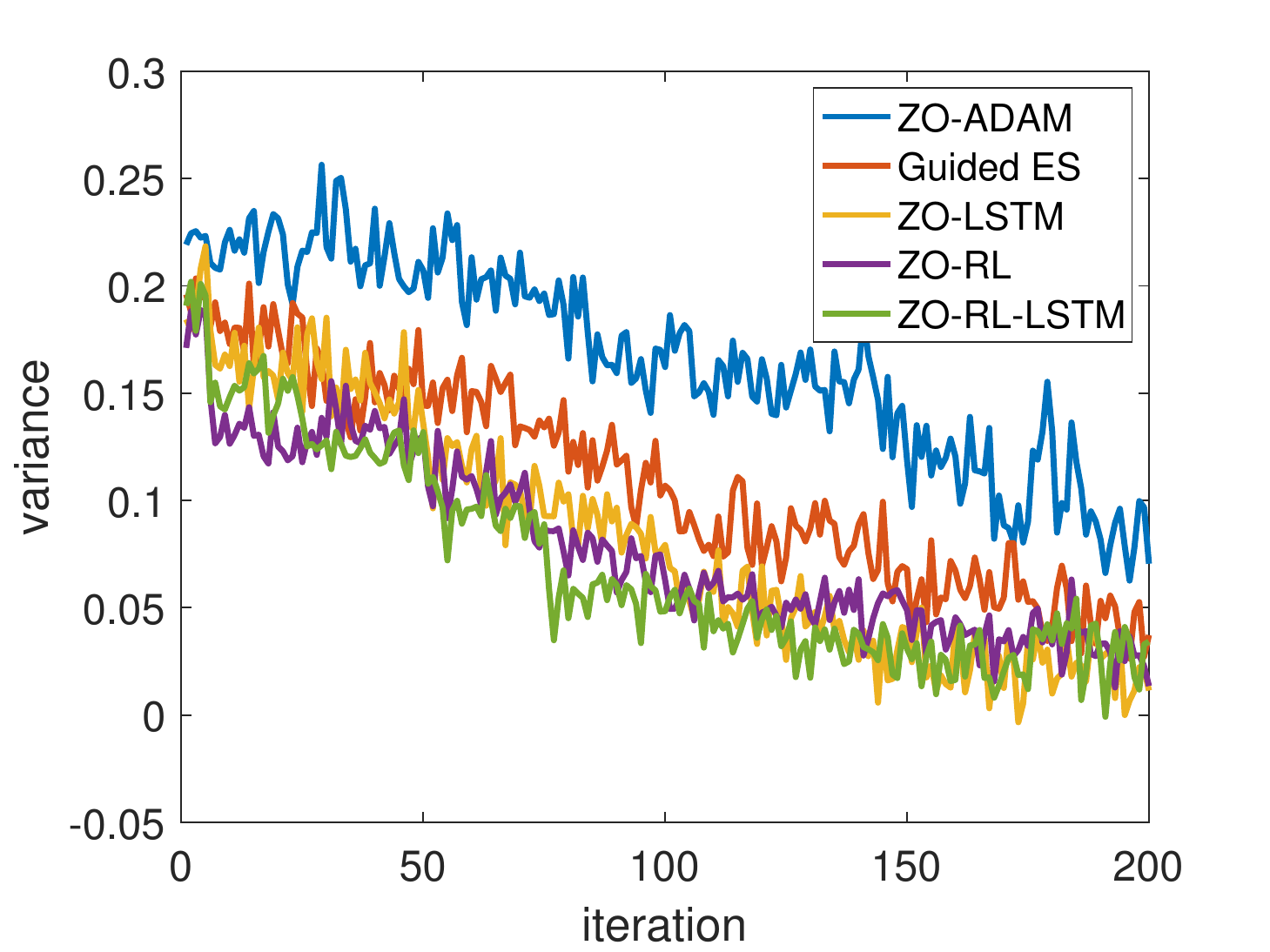}
		\caption{a9a}
	\end{subfigure}
	\caption{Non-convex optimization problems in the ADAM setting.}
	\label{fig:7}
\end{figure*}

\subsection{Discussion and Analysis}
Fig. \ref{fig:2} shows the black-box attack loss and variance versus iterations using different ZO optimizers in the SGD setting.
Fig. \ref{fig:3} visualizes the black-box attack loss and variance versus iterations using different ZO optimizers in the signSGD setting.
Fig. \ref{fig:4} plots the black-box attack loss and variance versus iterations using different ZO optimizers in the ADAM setting.
The loss curves are averaged over 10 independent random trails and the shaded areas indicate the standard deviation.
The results clearly show that our ZO-RL has significant advantage over random sampling to sample perturbed vectors, and outperforms the ZO algorithms that use the learned sampling distribution most of the time. 
This is due to the fact that our ZO-RL learn the sampling policy instead of random sampling that can reduce the variances of ZO gradient.

Fig. \ref{fig:5} draws the non-convex least squared loss and variance versus iterations using different ZO optimizers in the SGD setting.
Fig. \ref{fig:6} demonstrates the non-convex least squared loss and variance versus iterations using different ZO optimizers in the signSGD setting.
Fig. \ref{fig:7} illustrates the non-convex least squared loss and variance versus iterations using different ZO optimizers in the ADAM setting.
The loss curves are averaged over 10 independent random trails and the shaded areas indicate the standard deviation. 
Our ZO-RL has a definite advantage in sampling the perturbed vectors compared with random sampling, and our ZO-RL-LSTM can always obtain the best results by combining learned sampling policy and sampling distribution.
The results clearly demonstrate that our ZO-RL leads to much faster convergence and lower final loss under different parameter update settings compared to existing ZO algorithms.
The results also show that our ZO-RL can effectively reduce the variances of ZO gradient by learning the sampling policy that maximizes expected cumulative reward.

\section{Conclusion}

We proposed a new reinforcement learning based sampling policy for generating the perturbations in ZO optimization instead of using the existing random sampling. The learned sampling policy guides the perturbation (direction) in the parameter space to estimate a more accurate ZO gradient. To the best of our knowledge, our ZO-RL is the first algorithm to learn the sampling policy via reinforcement learning for ZO optimization which is parallel to the existing methods. Especially, our ZO-RL can be combined with the existing ZO algorithms that could further accelerate them. Experimental results of solving different ZO optimization problems show that our ZO-RL algorithm effectively reduces the variances of ZO gradient by learning the sampling policy, and converges faster than existing ZO algorithms in different scenarios.

\ifCLASSOPTIONcaptionsoff
  \newpage
\fi



\bibliographystyle{IEEEtran}
\bibliography{example_paper}
\end{document}